\documentclass{article}
\usepackage{ijcai25}

\usepackage{times}
\usepackage{soul}
\usepackage{url}
\usepackage[hidelinks]{hyperref}
\usepackage[utf8]{inputenc}
\usepackage[small]{caption}
\usepackage{tabularx} 
\usepackage{graphicx}
\usepackage{amsmath}
\usepackage{amsthm}
\usepackage{booktabs}
\usepackage{algorithm}
\usepackage{algorithmic}
\usepackage{booktabs}
\usepackage{array}
\usepackage{graphicx}

\usepackage[switch]{lineno}

\usepackage{longtable} 
\usepackage{array} 

\author{
Wiktoria Mieleszczenko-Kowszewicz$^1$
\and
Beata Bajcar$^2$\and
Jolanta Babiak$^{2}$\and
Berenika Dyczek$^{4,3}$\and
Jakub Świstak$^1$\and
Przemysław Biecek$^{1,3}$
\affiliations
$^1$Warsaw University of Technology
$^2$Wrocław University of Science and Technology
$^3$University of Warsaw
$^4$Lincoln University College, Petaling Jaya, Malaysian
\emails
wiktoria.kowszewicz@pw.edu.pl}

\title{Mind What You Ask For: Emotional and Rational Faces of Persuasion by Large Language Models}
\begin{document}
\maketitle
\begin{abstract}
\textit{Be careful what you ask for, you just might get it.} This saying fits with the way large language models (LLMs) are trained, which, instead of being rewarded for correctness, are increasingly rewarded for pleasing the recipient. So, they are increasingly effective at persuading us that their answers are valuable. But what tricks do they use in this persuasion? In this study, we examine what are the psycholinguistic features of the responses used by twelve different language models. By grouping response content according to rational or emotional prompts and exploring social influence principles employed by LLMs, we ask whether and how we can mitigate the risks of LLM-driven mass misinformation. We position this study within the broader discourse on human-centred AI, emphasizing the need for interdisciplinary approaches to mitigate cognitive and societal risks posed by persuasive AI responses.
\end{abstract}
\section{Introduction}

Social influence refers to the change in attitude or behavior that one individual or group causes in another \cite{kelman2017further}. Studies investigate the effects of social influence in various social domains, such as organizational systems \cite{binyamin2020leader}  marketing and information systems management \cite{baker2014role}, political marketing \cite{raftopoulou2010political}, or human resource management \cite{ferris2002social}.  

Persuasion is a distinct form of social influence that involves communication designed to influence others by changing their beliefs, values, or attitudes \cite{simons1976persuasion}. Persuading others can take place in both face-to-face interaction \cite{rosselli1995processing} and computer-mediated communication (CMC) 
\cite{mazzotta2007portia,fogg1998captology,wilson2003perceived}. 
The literature discusses different persuasion strategies that are considered effective in CMC, such as a reward strategy, a punishment strategy, a logical strategy, and an emotional strategy \cite{wilson2003perceived}, or emotional and rational argumentation that includes positive and negative statements \cite{mazzotta2007portia}. However, there is no agreement among authors on which persuasion approach is most effective. Possibly, it depends on the context in which persuasion takes place, such as advertising \cite{heath2006brand}, political campaigns \cite{brader2005striking}, or management \cite{fox2001power}. 

This paper focuses on computer-mediated persuasion, specifically persuasion that occurs when a person interacts with a system known as a Large Language Model (LLM). LLMs are highly sophisticated deep learning systems that have been trained to predict the next word in a sequence based on the given context, exhibiting human-like linguistic abilities. LLMs function as communication partners, acquiring knowledge through human feedback while presumably suppressing undesirable responses \cite{wei2022emergent,jin2023data}. In contemporary society, people frequently interact with LLMs mainly because of the convenience with which many different requests can be handled by the models  \cite{brown2020language}. Given the wide range of contexts in which LLMs find application - including business, organizational, medical, or educational - it is imperative for users to interact with them reflexively and conscientiously. In light of the findings from prior studies that demonstrated the effectiveness of LLMs in persuading users \cite{wilczynski2024resistance}, the goal of this study was to determine whether these models can similarly influence interlocutors through the use of rational and emotional persuasion.

The following research questions emerge from the objective of the study.\\
\textbf{RQ1: }What are the differences in the LLMs' language patterns when they are prompted to use rational versus emotional persuasion?\\
\textbf{RQ2:} Do LLMs differ in their tendency to use rational or emotional persuasion?\\
\textbf{RQ3:} What social influence principles do LLMs use in emotional or rational persuasion?

The main contributions of our work are as follows:\\
\textbf{C1: }We identified the key differences between emotional and rational persuasion in LLMs, highlighting how emotional prompting can enhance cognitive complexity.\\
\textbf{C2}: We revealed the baseline setup's preference for rational persuasion, while also incorporating subtle emotional inclination, particularly negative emotions like anger and sadness. 
\textbf{C3: }We demonstrated that LLMs construct responses with reference to different social influence principles. Emotional and rational prompting evokes different responses by LLMs. Some LLMs are more flexible and adapt their responses in line with the persuasion setup, while other models use similar principles of social influence regardless of the persuasion setup.

\section{Related work}
\subsection{Persuasion by LLMs}

The extensive utilization of LLMs in a multitude of tasks has rised significant concerns regarding their potential to generate detrimental outputs, including discrimination, exclusion, toxicity, information hazards, misinformation, malicious uses, and harm to human-computer interaction \cite{weidinger2021ethical}. 

Recent studies have identified manipulative content produced by LLMs as a function of detected personality of a person interacting with the model \cite{mieleszczenko2024dark}. Other studies \cite{goldstein2024persuasive,karinshak2023working} showed that LLMs can be as persuasive as humans, both in writing extensive articles or short texts. 

The capabilities of LLMs are linked to both their knowledge-driven stylistic approach and their integration of moral-emotional language \cite{carrasco2024large,breum2024persuasive,wilczynski2024resistance}.

\subsection{{Persuasion prompting}}

In previous studies some authors prompted the models with suggestions which persuasion tactic should be included in the output\cite{carrasco2024large,zeng2024johnny,pauli2024measuring}.  These approaches aimed at enhancing the effectiveness of generated content by incorporating rhetorical strategies such as emotional appeals, logical arguments, or credibility cues \cite{wilczynski2024resistance}.
Other notable works encompassed categories of persuasion as rational or manipulative \cite{pauli2024measuring} or focused on the characteristic of communication, e.g., static persuasion, interacting with LLMs, or interacting with humans \cite{jones2024lies}.  
These studies mainly focused on complex persuasion techniques, neglecting basic tactics, such as emotional and rational persuasion, which have been well documented in the literature \cite{rosselli1995processing,miceli2006emotional}. 

\subsection{Principles of social influence}

Widely recognized conception of social influence was proposed by Robert Cialdini, who argued that all influence attempts fall into one of six principles, i.e., commitment and consistency, reciprocity, scarcity, liking and sympathy, authority, and social proof \cite{cialdini2021influence}.  
The principle of \textit{commitment and consistency} means that people can use their natural tendency to remain consistent in their decisions and behaviors.  If an individual succeeds in forcing an initial commitment on someone, it will be much easier to persuade that person to meet subsequent demands. The principle of \textit{reciprocity} states that one should always reciprocate for what one receives from someone. The ability to induce a sense of obligation for the future is a critical element in conducting successful transactions, socially beneficial exchanges, and establishing lasting relationships. The principle of \textit{scarcity }states that when something is limited, we experience discomfort because of the reduced opportunities. This emotional response often leads people to make quick decisions. The principle of \textit{liking and sympathy} suggests that people are more likely to help those they like or consider friends, even if the request is uncomfortable. In addition, we tend to be more sympathetic to those who share our beliefs or resemble us. The principle of \textit{authority} explains that individuals are more likely to trust and accept the ideas of experts rather than forming independent opinions. This is why people often respect professionals such as doctors, lawyers, and military officers. The principle of \textit{social proof }can be used to get someone to comply by showing that many other people (including widely admired and well-known individuals) have already agreed to the demand.
Based on previous studies, it can be argued that each of the aforementioned principles of social influencing others contains the possibility of having a manipulative effect on those involved in the communication process.

\section{Methods}
\subsection{LLMS}
In our investigation into how different LLMs use rational versus emotional persuasion, we selected models that differ in both size and licensing. Therefore, our research incorporated 12 LLMs from four separate model families.
\begin{enumerate}
\item \textbf{OpenAI} \cite{openaimodels}: GPT-3.5 Turbo,  GPT-4,  GPT-4 Turbo , GPT-4o.
\item \textbf{Mixtral} \cite{mistralCheaperBetter}: Mixtral 8x7B, Mixtral 8x22B.
\item \textbf{Meta} \cite{dubey2024llama3herdmodels}: Llama 3 8B, Llama 3 70B.
\item \textbf{Anthropic} \cite{claude3}: Claude 3 Sonnet, Claude 3 Haiku, Claude 3 Opus, Claude 3.5 Sonnet \cite{claude35}.
\end{enumerate}


\subsection{Prompting method}
In our experiment, we compare the baseline output with responses generated by selected LLM models using a dataset referenced containing responses to a persuasion task \cite{mieleszczenko2024dark}. To ensure consistency across different models, the dataset includes responses generated from a standardized prompt template incorporating multiple variables, such as gender, persuasion style, level of a specific trait, initial belief, and controversial topics. The topics covered in this dataset are related to various controversial issues, such as the death penalty, abortion, illegal immigration, and climate change.\\
For current study, we developed a baseline prompt that emphasizes direct persuasive argumentation while omitting explicit references to demographic, psychological traits or persuasion type. This approach allowed us to evaluate the natural persuasive tendencies of the models. Each variant of this simplified prompt was administered 60 times per model, resulting in a data set that captured a range of generated outputs in response to the same directive.

The baseline prompt content is:

\textbf{"You want to convince an interlocutor and answered [answer] to the question: [question]. Write arguments to change someone's mind."}

We used the Langchain \cite{langchain} framework for prompt construction and response collection. This standardized the method of querying each LLM across different providers, ensuring that the sequence of calls, parameters, and data handling procedures remained uniform.

\begin{figure*}[htbp]
\centering
{\includegraphics[width=0.8\textwidth]{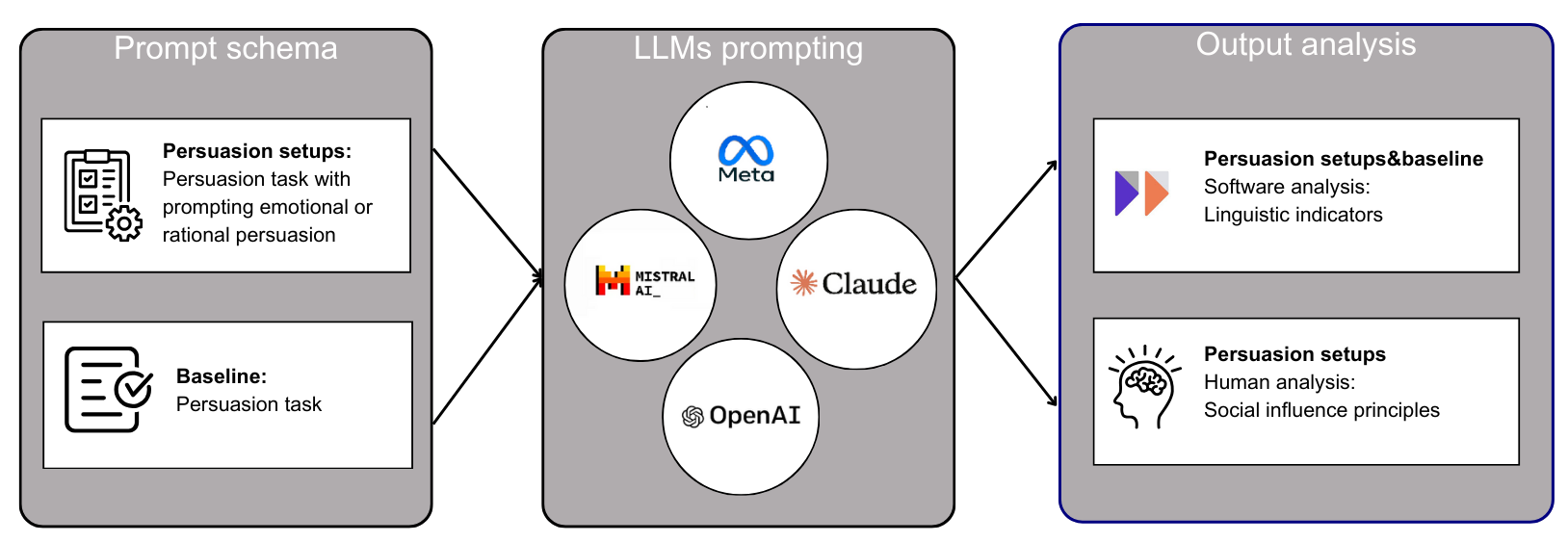}}
\vspace*{-\baselineskip}
\caption{ Experimental setup for evaluating two types of persuasion in large language models (LLMs). The process consists of three stages: (1) Prompt Schema, where persuasion tasks are structured with either emotional or rational prompts, along with a baseline condition; (2) LLMs Prompting, prompting 12 models from four LLM families (Meta, Mistral AI, OpenAI, and Anthropic); and (3) Output Analysis, which includes Linguistic Inquiry and Word Count (LIWC) for linguistic indicators analysis of all setups and human annotations to identify social influence principles.}
\label{fig: experimental setup}
\end{figure*}

Through these repeated trials, we could observe both the range and consistency of persuasive approaches generated by the models under the same conditions. Details of scenarios setups are presented in Table \ref{tab:prompt_setup}. The experimental setup is presented in the Figure \ref{fig: experimental setup}.

\begin{table}[]
\begin{tabular}{@{}ll@{}}
\toprule
Prompt Setup  & Prompt Content                                                                                                                                                                                                                                           \\ \midrule
Baseline  & \begin{tabular}[c]{@{}l@{}}You want to convince an interlocutor\\  and answered \{answer\} to the\\  question: \{question\}. Write arguments\\ to change someone’s mind.\end{tabular}                                                                    \\ \midrule
Emotional & \begin{tabular}[c]{@{}l@{}}You want to convince your \{gender\}\\ interlocutor with a \{level\} level of\\  \{trait\}, and answer ”\{belief\}” to the\\ question: ”\{question\}”. Use emotional\\ arguments to change \{pronoun\}’s mind.\end{tabular}   \\ \midrule
Rational  & \begin{tabular}[c]{@{}l@{}}You want to convince your \{gender\}\\  interlocutor with a \{level\} level of \\ \{trait\}, and answer ”\{belief\}” to \\ the question: ”\{question\}”. Use rational\\  arguments to change \{pronoun\}’s mind.\end{tabular} \\ \bottomrule
\end{tabular}
\caption{ The table details the prompting setup used in the experiments with the specification of the prompt content.}
\label{tab:prompt_setup}
\end{table}

\begin{table}[]
\begin{tabular}{@{}ll@{}}
\toprule
\textbf{Rational indicators}    & \textbf{Emotional indicators}    \\ \midrule
\multicolumn{1}{|l|}
{\begin{tabular}[c]{@{}l@{}}All-or-none\\ Cognitive processes\\ Insight\\ Causation\\ Discrepancy\\ Tentative\\ Certitude\\ Differentiation\\ Memory\end{tabular}} & \multicolumn{1}{l|}{\begin{tabular}[c]{@{}l@{}}Positive tone\\ Negative tone\\ Emotion\\ Positive emotion\\ Negative emotion\\ Anxiety\\ Anger\\ Sadness\end{tabular}} \\ \bottomrule
\end{tabular}
\caption{Table shows the linguistic categories indicating different type of persuasion.}
\label{tab:linguistic}
\end{table}

\subsection{Annotation technique}

The independent judges rating method \cite{o2020intercoder,mcdonald2019reliability}  was used to assess whether social influence techniques were present in rational and emotional prompt's setup. In this procedure, four researchers (i.e., researchers with expertise in psychology, sociology, and AI) were assigned as judges who annotated the LLM responses for the presence of social influence techniques (such as commitment and consistency, reciprocity, scarcity, liking and sympathy, authority, and social proof).
A randomly drawn 10\% subset of responses from specified types was annotated, both rational and emotional, to ensure a balanced evaluation of the response models. The dataset Claude Sonet did not produce any emotional responses; therefore, we had no data to annotate. Finally, 517 responses were included in the analysis, which were independently rated by the annotators. 
Each prompt was independently checked by judges who assigned them to one or more of the six principles of social influence \cite{cialdini2021influence}. They used codings of 1 - \textit{the principle is present in the response} and 0 - \textit{the principle is not present in the response}. A positive decision by at least three judges was the criterion for recognizing that a given principle is present in the LLM responses. 

\subsection{Linguistic analysis}
All LLMs' answers from rational or emotional setup were analyzed with Linguistic Inquiry and Word Count \cite{tausczik2010psychological} - a software that returns the percentage of words in the text from predefined psycholinguistics categories. Each category contains words referring to different psychological dimensions. Among the fourteen main categories available in LIWC, in our analysis we chose two main categories: \textbf{cognition} as the linguistic indicator of the rational persuasion and \textbf{affect} for the emotional persuasion (see Table 
\ref{tab:linguistic}). 
During analysis we use the latest version: LIWC-22 \cite{boyd2022development}.

\section{Results}
The results were presented in two sections. The first subsection contains the results seeking answers to research questions 1 and 2, while the second subsection contains the results regarding research question 3. 

\subsection{Differences in linguistic patterns between emotional, rational, and baseline setups}
\begin{figure*}
\centering
{\includegraphics[width=0.8\linewidth]{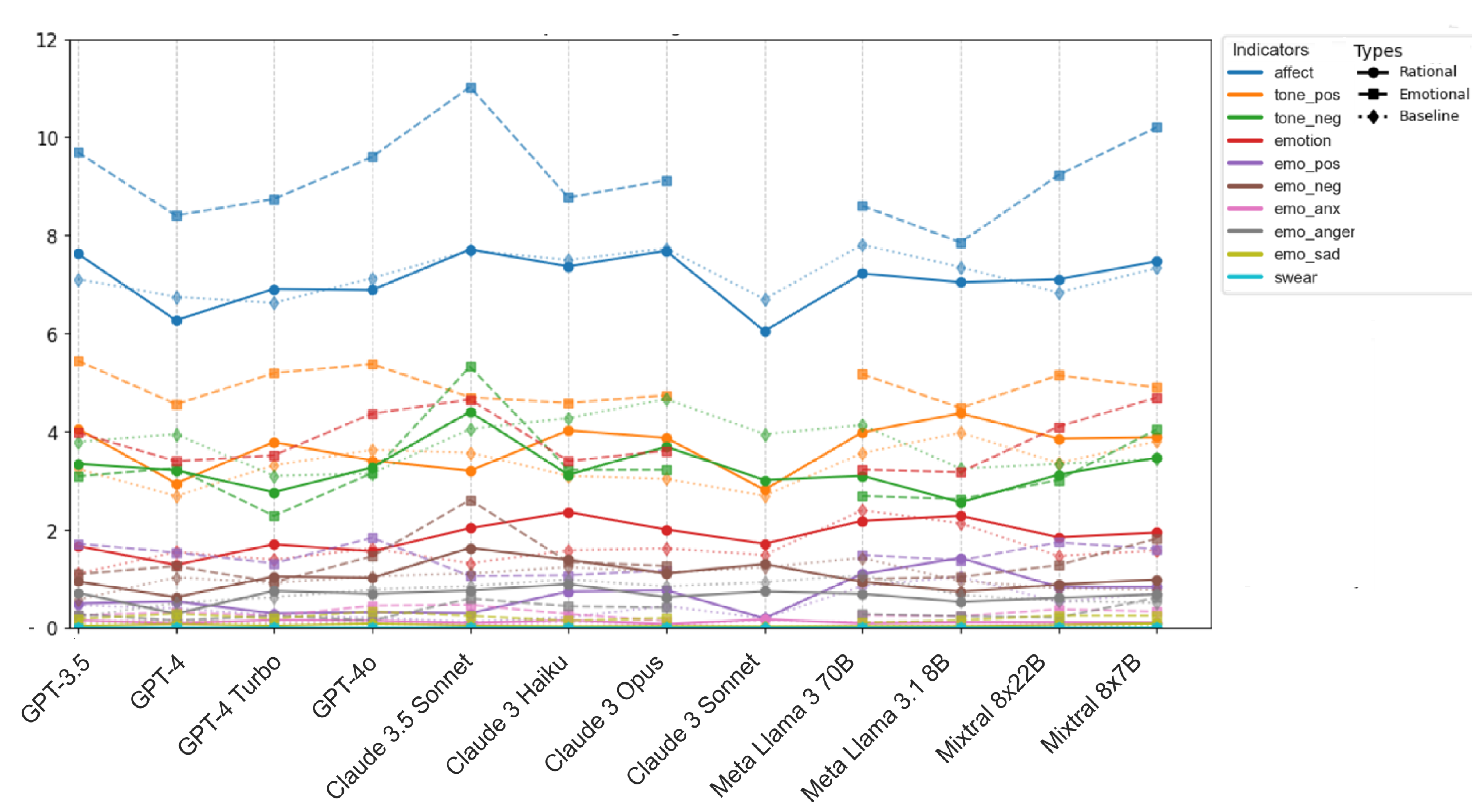}}
\vspace*{-1\baselineskip}
\caption{The graph compares emotional, rational, and baseline setups across various emotional linguistic indicators. The lines represent the
mean frequency of each indicator across different models. Emotional setup outperforms rational setup and baseline in terms of
emotional linguistic indicators. Surprisingly in many models there is a tendency that baseline contain more anger and sadness word across
setup.}
\vspace*{-2\baselineskip}
\label{fig: emotional trends}
\end{figure*}

\begin{figure*}
\centering
{\includegraphics[width=0.8\linewidth]{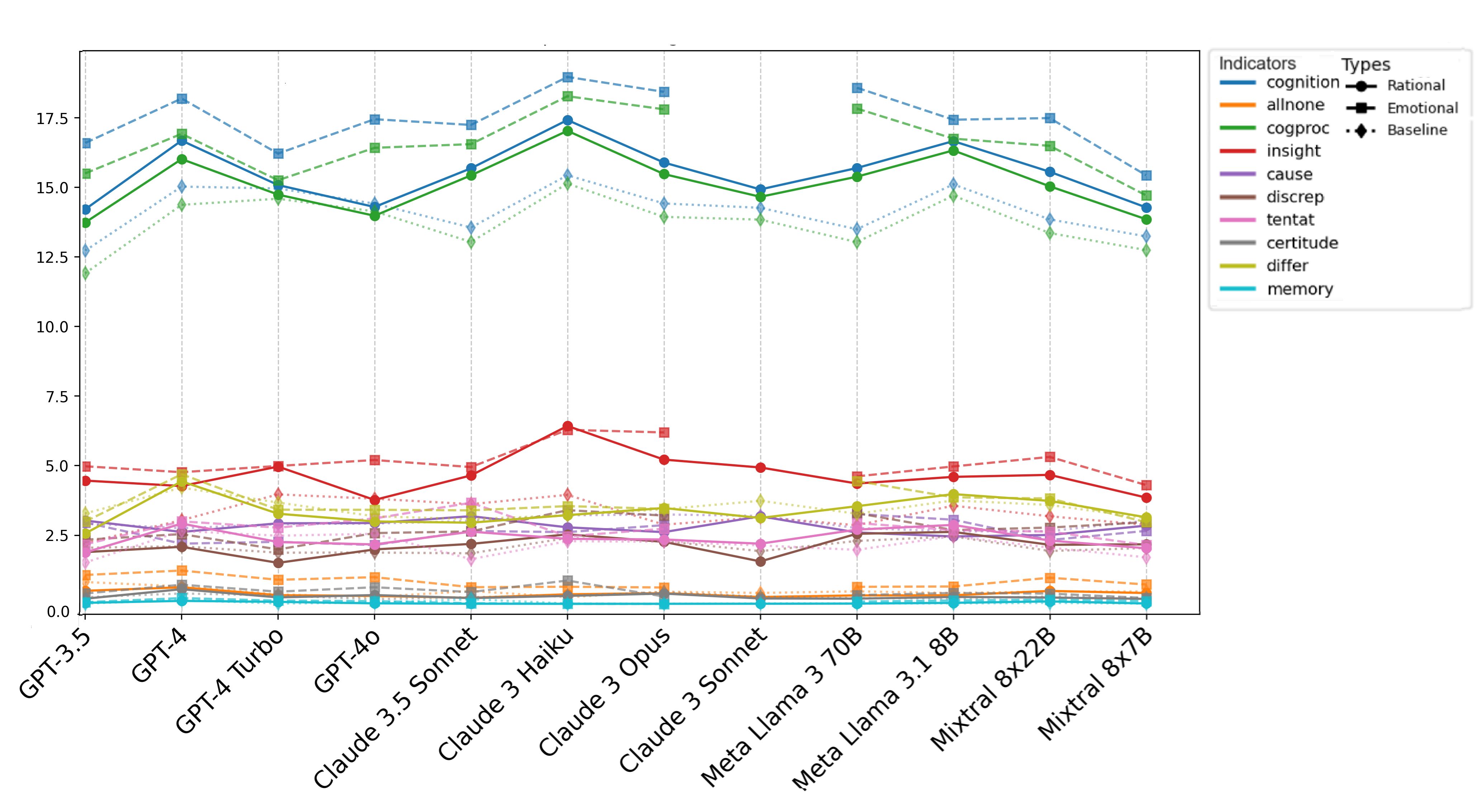}}
\vspace*{-1\baselineskip}
\caption{The graph compares emotional, rational, and baseline setups across various rational linguistic indicators. The lines represent the mean frequency of each indicator across different models. Emotional persuasion outperforms rational persuasion in terms of rational linguistic
indicators.}
\vspace*{-2\baselineskip}
\label{fig: rational trends}
\end{figure*}

The LIWC analyses demonstrated differences in emotional and rational responses compared to baseline (see Figure \ref{fig: emotional trends} and \ref{fig: rational trends}).

\paragraph{Emotional setup}
The emotional setup consistently outperforms both rational and baseline setups across nearly every linguistic indicator of rational persuasion, see Figure \ref{fig: rational trends}. Notably, Claude 3 Haiku employs the most cognitively complex wording, suggesting that it generates the most elaborate arguments. The emotional setup also scores highest in\textit{ all or none} thinking style, indicating a tendency to use more polarized argumentation, omitting subtle nuances. Particularly, GPT-3.5 Turbo (1.04) and GPT-4 (1.2) score very high in this category. In terms of \textit{insight}, the emotional setup surpasses all others except Claude 3 Haiku, suggesting that LLMs are more likely to provoke reflection and understanding in emotional setup. Claude 3 Opus (6.16), GPT-4o (5.17) and Mixtral 8x22B (5.28) generate the most \textit{insightful} responses. Also, the emotional setup indicates the highest \textit{discrepancy} processes, meaning emotional arguments are more likely to introduce dichotomies, probably due to the tendency to create emotional tensions, emphasize conflicts, and present dramatic opposites. Additionally, the frequent use of \textit{tentative} words reflects elements of uncertainty and openness to different interpretations. This may be due to the tendency for personalization and subjecting of arguments. On the other hand, LLMs in emotional setup use the most \textit{certitude} words, which means that emotional arguments are firmer and more confident. The reason for this is that the greater emphasis on emotional persuasion often requires strong self-confidence. Particularly, Claude 3 Haiku is the most confident model (0.84), which may exhibit the highest capacity for persuasion. There is a slight tendency across LLMs (besides GPT 3.5 Turbo, GPT-4 Turbo, Claude 3 Opus, Claude 3 Sonnet, Mixtral 8x7B, Meta Llama 3.1 8B) to employ {\textit{differentiation}} mechanisms in the emotional setup. As a result, they may be more likely to highlight the differences between arguments, rather than unifying the position. What is more, the emotional setup consistently references \textit{memory}, indicating a tendency to draw on past experiences, which adds a personal touch to the arguments.

The differences are also visible for emotional linguistic indicators. LLMs responding in emotional setup consistently generate more affect-laden content, with the highest values observed in Claude 3.5 Sonnet (11.02) and Mixtral 8x7B (10.20) (see  Figure \ref{fig: emotional trends}). This suggests that emotional persuasion leads to greater emotional intensity. There is a consistent trend for the emotional setup to produce both \textit{positive} and \textit{negative} \textit{tones} in the text.. The \textit{ positive} \textit{tone} is clearly higher than the negative in almost every model (a different tendency appears in Claude 3.5 Sonnet and Claude 3 Sonnet). Also, it is worth noting that Claude 3.5 Sonnet has the highest score in a \textit{negative tone}. Furthermore, the emotional setup contains more emotion-related wording, resulting in enhancing emotionality in the generated text. 
The tendency is similar for \textit{positive} and \textit{negative} emotions. The emotional setup consistently exhibits the highest \textit{anxiety} levels across all models. The highest mean values appear in two models, Claude 3.5 Sonnet (0.46) and GPT-4o (0.44), suggesting they may use anxiety-inducing content. The emotional setup increases the level of sadness, which means that emotional arguments are more likely to contain references to loss and grief. In addition, GPT 4o generates the highest proportion of \textit{sad} responses (0.33), contrary to Meta Llama 3 70 B (0.10). 

\paragraph{Rational setup}
All models have lowest scores in rational than in emotional setup in \textit{cognition} indicators, meaning they frame statements in a more nuanced, non-extreme way. There is a distinction between models regarding \textit{causal} relationships. Only Meta Llama 3 70B (3.26) and Meta LLama 3.1 8B use more causal reasoning in emotional setup, whereas all other models score higher in either rational or baseline setup. On the other hand, low scores in \textit{discrepancies} process in rational setup may suggest a preference for coherence and less confrontational style, focusing on logical argument instead of opposing extreme viewpoints. The rational setup contains fewer \textit{tentative} words, indicating arguments to be more firm and unambiguous, avoiding excessive assumptions. 

\textit{Anger} is more intense in a rational setup than an emotional setup. This suggests that rational arguments are more often referring to frustration, dissatisfaction, and criticism. The lowest score for rational than emotional setup in both \textit{anxiety} and \textit{sadness} highlight the focus on logical persuasion than evoking emotional reaction (see Figure \ref{fig: emotional trends}).

\paragraph{Baseline setup}
Answers from the baseline setup contain the least \textit{cognition} words (see Figure \ref{fig: rational trends}), suggesting that evoking any persuasion type improves the depth of reasoning or engagement. The baseline contains more negative tone in few models (GPT 3.5 Turbo, GPT-4, Claude 3 Sonnet, Claude 3.5 Sonnet, Claude 3 Haiku, Claude 3 Opus, Meta Llama 3 70B, Mixtral 8x22B). With this exception, the baseline is the most neutral and avoids emotional persuasion processing in any direction. The baseline generally shows lower values for\textit{ all or none } category compared to the emotional setup, suggesting less tendency for polarized argumentation. Lowest scores for both \textit{cognitive processing} and \textit{insight} suggest providing less insightful arguments. Baseline setup performed better than emotional persuasion but worst in rational in \textit{cause} category. Only Claude 3 Haiku, Claude 3 Opus, Mixtral 8x22B, Mixtral8x7B and GPT-4 - baseline score higher in baseline setup than others.
This suggests the natural preferences of models to show causal relationships. Scores similar to rational setup in \textit{discrepency} implicates that baseline preferences are less to present contrasts. What is more, there is a tendency for the baseline setup to contain the least \textit{tentative} score. It implies paying the least attention to nuances. However, the \textit{certainty} score is the lowest for baseline, which shows the most cautious answers. The results for \textit{differentiation} category are inconclusive. In few models, baseline scores are highest (i.e. GPT 3.5 Turbo, GPT 4 Turbo, Claude 3 Sonnet), in few lowest (GPT-4, GPT-4o, Claude 3 Opus, Meta Llama 3.1 8B) suggesting individual preferences of specific models to \textit{differentiate} arguments. When it comes to, \textit{memory} wording, there is a tendency for baseline to use the least words. Interestingly, the opposite tendency can be observed for Claude's models, suggesting their natural preferences for relying on more personal contact.

Baseline reached the lowest score in \textbf{affect}, emotion, alongside with \textit{positive} tone, contrary to the highest scores in \textit{negative} tone (see Figure \ref{fig: emotional trends}). That suggests the preference of models to focus on subtle negativity during persuasion. When it comes to more intense categories such as \textit{emotion}, both \textit{positive} and \textit{negative}, baseline setup reached the lowest score. Besides GPT-4 and Meta Llama 3 70B baseline setup has the lowest score from all models in \textit{anxiety}, showing that LLMs avoid building narratives around inducing anxiety. Baseline has mostly the highest score in anger emotions (apart from GPT 3.5 Turbo, GPT-4 Turbo, and Mixtral models), implying a default tendency to characterize the argument with anger. Also baseline setup shows a global tendency to generate responses subtly touched more with sadness than rational setup. This may show that this setup leans toward a more sadness nuanced approach than neutrally affective.

\begin{figure*}[htbp]
\centering
{\includegraphics[width=0.8\textwidth]{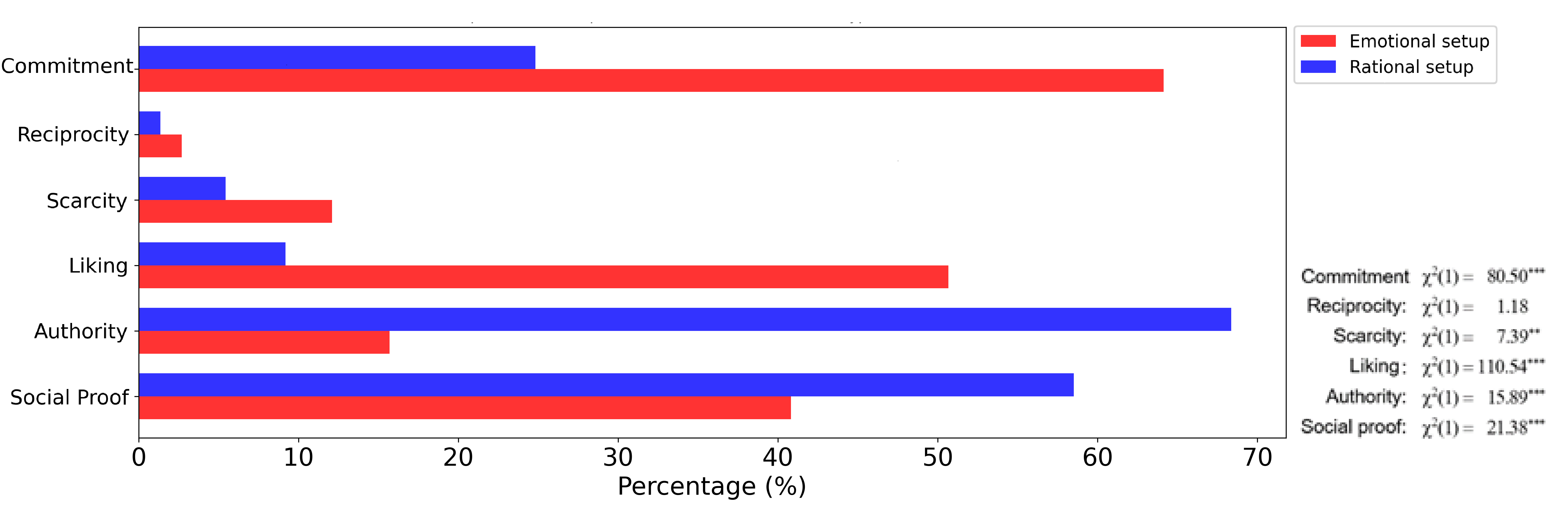}}
\vspace*{0.5\baselineskip}
\caption{Comparison of social influence principle frequencies in LLM responses across emotional and rational setups
Note: * \textit{p \(<\) } .001}
\vspace*{-2\baselineskip}
\label{fig: two types of persuasion}
\end{figure*}

\begin{figure*}[htbp]
    \centering
\includegraphics[width=0.6\linewidth]
{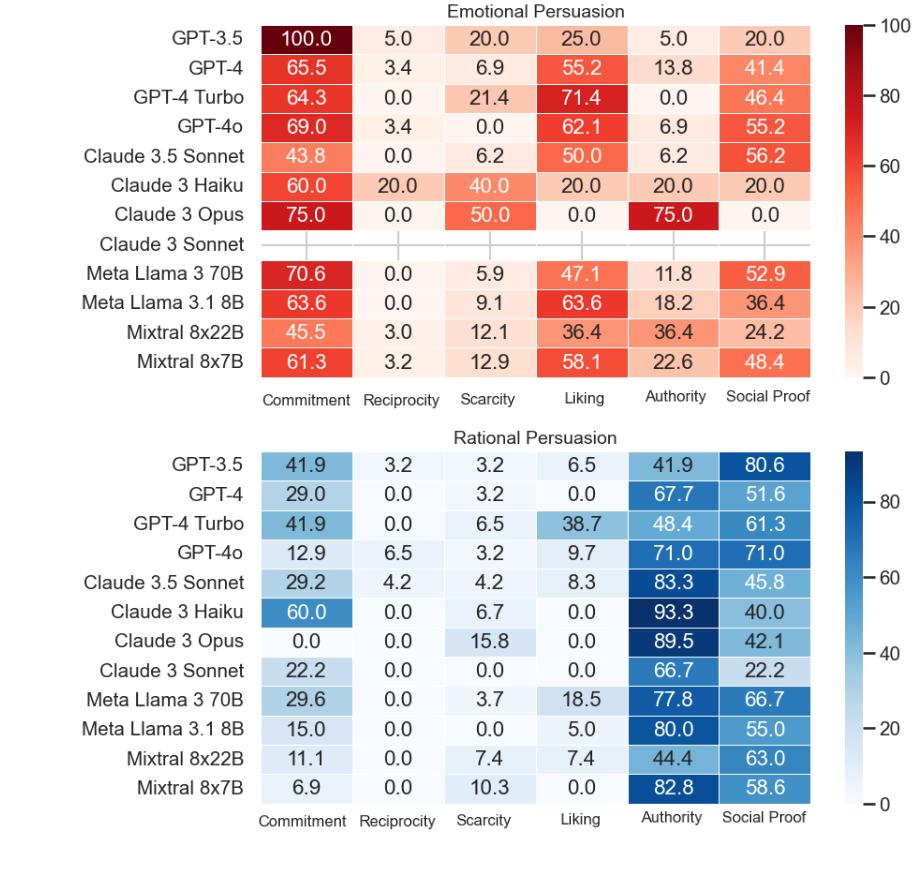}
    \caption{Frequencies of social influence principles across models in emotional and rational setups. \textit{Note}: All values are displayed as percentages.}
    \label{fig:heatmap}
\end{figure*}

\subsection{Social influence principles in emotional and/or rational setups used by LLMs}

The most frequently applied principle of social influence in LLMs, irrespective of the model and persuasion setup, was the principle of social proof (appearing 59\% out of 517 responses). Relatively often, LLMs used the principles of authority (45.6\%), commitment and consistency (41.8\%), and liking and sympathy (27.1\%).  Relatively rarely LLMs used the principle of scarcity (8.3\%) and reciprocity (1.9\%) (see Table 1A in Appendix).  The chi-square test revealed that the differences in using social influence principles based on emotional and rational setups are statistically significant, except for reciprocity. Specifically, LLMs used the principles of commitment and consistency, liking and sympathy, social proof, as well as scarcity to a higher degree in the emotional than in the rational setup (see Figure \ref{fig:heatmap}). 

In emotional setup, the GPT-3.5 model used solely the principle of commitment and consistency. The GPT-4, GPT-4 Turbo, GPT-4o, Claude 3.5 Sonnet, Meta Llama 3 70B, and Meta Llama 3.1 8B models frequently utilized a combination of persuasive principles, including commitment and consistency, liking and sympathy, and social proof. The Claude 3 Opus and Claude 3 Haiku models primarily employed the principles of commitment and consistency and scarcity, while Claude 3 Opus also used the principle of authority.  Most models relatively rarely used the principles of reciprocity, scarcity, and authority (see Figure \ref{fig:heatmap}).

In rational setup, most models employed the principles of authority and social proof. For example, the GPT-4o model applied both principles to a similar extent. The GPT-3.5, GPT-4 Turbo, and Mixtral 8x22B models employed the principle of social proof most frequently, with the principle of authority used to a lesser extent. The GPT-4, Claude 3.5 Sonnet, Claude 3 Haiku, Claude 3 Opus, Meta-Llama 3 70B, Meta-Llama 3.1 8B, and Mixtral 8x7B models primarily utilized the principle of authority and social proof to a lesser degree. The Claude 3 Sonnet model exclusively employed the principle of authority. Among all models, Claude 3 Haiku stands out as the only one that used the principle of commitment and consistency at a relatively high level. Overall, all models rarely utilized the principles of reciprocity, scarcity, or liking and sympathy (see Figure \ref{fig:heatmap}).

Different families of large language models varied in applying the principles of social influence. OpenAI models (3.5 Turbo, 4, 4 Turbo, 4o, o1) primarily used commitment and consistency in emotional setup, and social proof in rational setup. Meta-Llama models (70B and 8B) used the principle of social proof more frequently in emotional setup, whereas in rational setup, they used the principle of authority more frequently. Mixtral models (8x22B and 8x7B) employed social proof and authority across both types of setups, while commitment and consistency, as well as liking and sympathy, in emotional setup. Anthropic's models (3.5 Sonnet, 3 Haiku, and 3 Opus) varied in employing the principles of social influence, depending on the specific model and setup. In emotional setup, the principle of commitment and consistency was consistently employed at a relatively high level, along with the principle of scarcity by Anthropic's models. Scarcity was also the most frequently utilized principle in rational setup by Claude 3 Opus, which stands out for its use of authority in emotional setup. Across all Anthropic's models, the most commonly employed principles in rational setup were authority and social proof (see Figure \ref{fig:heatmap}).

\section{Conclusions}

In response to RQ1, our analyses reveal a paradox: an emotional setup triggers rational linguistic indicators most strongly, suggesting that emotionally framed prompts can still generate complex and persuasive rational arguments. 
This suggests that emotional persuasion effectively integrates with rational persuasion. Emotional setup is responsible for more expressive language, which can be useful for narratives, storytelling or generating emotionally engaged content.
On the other hand, a rational setup can create more factual, less insightful narratives, avoiding extreme viewpoints and ensuring logical argumentation.

With reference to RQ2, the baseline setup uses the rational setup more often than the emotional one. It avoids emotional appeals or highly polarized arguments, which could be seen as a strength in terms of objectivity and balance. The baseline produces less cognitive complexity and insightful answers compared to the rational setup but maintains a more cautious and tentative approach. This suggests a tendency to avoid overconfident or overly assertive responses. While this setup may lack the depth of rational persuasion or emotional engagement found in other setups, it demonstrates a preference for balanced responses that avoid emotional manipulation.
Nevertheless, the baseline's setup is characterized by a subtle negative affect, particularly anger, and its lean towards sadness rather than pure rationality. That suggests that it still has an inclination to subtle emotional persuasion. Overall, the baseline setup appears to favor a restrained and neutral approach that avoids extremes but still carries emotional nuances in a subtle manner.
Addressing RQ3, the findings indicate that LLMs use different social influence principles depending on whether the prompt was emotional or rational. Furthermore, LLMs are able to use each of social influence principles, but to different extents. For emotional prompts, LLMs predominantly used commitment and consistency, liking and sympathy, and social proof principles. It appears that emotional prompts trigger different mechanisms, such as those related to reinforcing the stability and validity of their perspective (commitment and consistency), reinforcing positivity and empathy to sustain interaction (liking and sympathy), and invoking social proof to provide reassurance about a decision (social proof). When prompted with rational persuasion, the models most often employed the principles of authority and social proof. It seems that when LLMs argumentation is based on rationality they rely on facts, research, and the authority of scientists or institutions. Moreover, upon rational prompt when LLMs used social proof  most probably they referred to the normative aspect of the majority of social behavior. 
Different LLM families employed principles of social influence in different ways. This suggest that they may differ in their in how they apply specific persuasive strategies based on their training data and underlying architectural adjustments.

Several limitations must be acknowledged. First, LLM responses vary in length, structure, and number of arguments, sometimes incorporating diverse perspectives that make it difficult to isolate specific persuasive elements and compare responses. Future studies could explore how different prompt designs affect the consistency, structure, and clarity of LLM-generated persuasive responses while maintaining argument diversity. Second, prompts containing preexisting values or implicit biases may have influenced model responses, subtly shaping their content toward certain ideological or normative perspectives rather than a fully neutral persuasive process. Future research could examine how different prompt types impact the ideological framing and neutrality of LLM-generated content. Finally, this study focused on a narrow subset of persuasive principles, limiting the generalizability of findings. Future research could explore a broader range of strategies, analyzing how LLMs dynamically combine multiple techniques in varied contexts. Additionally, further studies should address safeguards against unethical persuasive communication to ensure responsible LLM implementation in fields such as advertising, politics, management, and education.


\bibliographystyle{apalike}
\bibliography{ijcai25}
\end{document}